\documentclass{article}

\usepackage[english]{babel}

\usepackage[letterpaper,top=2cm,bottom=2cm,left=3cm,right=3cm,marginparwidth=1.75cm]{geometry}

\usepackage{setspace}
\usepackage[utf8]{inputenc} 
\usepackage[T1]{fontenc}    
\usepackage{url}            
\usepackage{booktabs}       
\usepackage{amsmath}
\usepackage{amsfonts}       
\usepackage{nicefrac}       
\usepackage{xcolor}         
\usepackage{amssymb}
\usepackage{tikz}
\usepackage{comment}
\usepackage{pifont}
\usepackage{bbm}
\usepackage{multirow}
\usepackage{rotating}
\usepackage{graphicx}
\usepackage{colortbl}       
\usepackage{wrapfig}
\usepackage[colorlinks=true, allcolors={blue!55!black}]{hyperref}
\usepackage{authblk}
\usepackage[numbers,sort&compress]{natbib}
\usepackage[final]{microtype}

\renewenvironment{abstract}{%
  \small
  \begin{center}{\bfseries \abstractname\vspace{-.5em}}\end{center}%
  \quote
}{%
  \endquote
}

\ifdefined\CBox\else\newsavebox\CBox\fi
\providecommand{\fakebf}[1]{\sbox\CBox{#1}\resizebox{\wd\CBox}{\ht\CBox}{\textbf{#1}}}
\providecommand{\uln}[1]{\smash[b]{\ensuremath{\underline{#1}}}}
\providecommand{\rowstrutc}{\rule[-2.6pt]{0pt}{10.2pt}}
\providecommand{\headstrut}{\rule[-4pt]{0pt}{64.5pt}}
\providecommand{\sepm}[1]{{\scriptsize$\,\pm#1$}}
\providecommand{\val}[2]{$#1$\sepm{#2}}
\providecommand{\valb}[2]{\fakebf{#1}\sepm{#2}}
\providecommand{\valu}[2]{\uln{#1}\sepm{#2}}
\providecommand{\sv}[1]{$#1$}
\providecommand{\svb}[1]{\fakebf{#1}}
\providecommand{\svu}[1]{\uln{#1}}

\date{}

\title{\textbf{In-Context Multiple Instance Learning}}
\renewcommand*{\thefootnote}{\fnsymbol{footnote}}

\author[1,2,3,*]{Alexander M\"ollers}
\author[1,2,3,*]{Marvin Sextro}
\author[1,2]{Julius Hense}
\author[1,2,4,$\dagger$]{\authorcr Gabriel Dernbach}
\author[1,2,5,6,$\dagger$]{Klaus-Robert M\"uller}
\affil[1]{Berlin Institute for the Foundations of Learning and Data, Berlin, Germany}
\affil[2]{Machine Learning Group, Technische Universit\"at Berlin, Berlin, Germany}
\affil[3]{Aignostics, Berlin, Germany}
\affil[4]{Institute of Pathology, Charit\'e -- Universit\"atsmedizin Berlin, Berlin, Germany}
\affil[5]{Max-Planck Institute for Informatics, Saarbr\"ucken, Germany}
\affil[6]{Department of Artificial Intelligence, Korea University, Seoul, Korea}
\begin{document}
\maketitle
\renewcommand\thefootnote{*}\footnotetext{Joint first authors}
\renewcommand\thefootnote{$\dagger$}\footnotetext{Corresponding authors: \texttt{gabriel.dernbach@gmail.com}, \texttt{klaus-robert.mueller@tu-berlin.de}}

\begin{abstract}
Multiple Instance Learning (MIL) addresses problems where supervision is available at the level of bags of instances and has been successfully applied in fields ranging from computational pathology to satellite imagery. Nevertheless, existing algorithms struggle in the low-label regime that characterizes many real-world applications. Flexible models overfit and rigid ones fail to adapt to the task at hand. We show that pretraining an in-context learner with a Perceiver-style architecture on synthetic data yields a model that can solve new tasks from a handful of labeled bags. At inference time, classification happens in a single forward pass and requires no gradient updates. We propose and investigate different synthetic data generators for bag-structured data and find that they capture complementary inductive biases. A model pretrained on a mixture of these generators inherits their per-task strengths and achieves the best average performance across twelve MIL benchmarks, outperforming supervised baselines that require task-specific training.
\end{abstract}

\section{Introduction} \label{sec:intro}

Multiple Instance Learning (MIL) is a widely used framework for problems where labels are available only at the level of collections (\emph{bags}) of instances rather than for individual ones. Applications of MIL span computational pathology, drug activity prediction, satellite imagery, and text classification \cite{otsu_multiresolution_2023, waqas_exploring_2024,idaji_beyond_2026,jeong_scmild_2026}. A persistent challenge in many of these domains is label scarcity, because real-world MIL datasets often contain only a few dozen labeled bags \cite{dietterich_solving_1997, babak_diagnostic_2017}. Supervised training of a robust aggregation function on such limited data is fragile as flexible models such as attention-based MIL \cite{ilse_attention_2018} and TransMIL \cite{shao_transmil_2021} may overfit, while restrictive models encode strong inductive biases that may not match the task at hand~\cite{mi_svm_andrews_2002}. 

Supervised MIL pretraining \cite{shao_do_2025} can mitigate this for applications with access to large labeled source corpora with similar examples, but many applications lack such resources and transfer is not always guaranteed. Self-supervised methods that learn bag-level representations~\cite{xu_whole-slide_2024, ding_multimodal_2025} remove the need for pretraining labels but must compress instance information into a fixed representation without knowledge of the downstream task, potentially discarding features that are critical for a specific problem. Thus, so far no method enables learning from a small number of labeled bags at inference time without also requiring a large task-relevant pretraining corpus. 

\begin{figure}[t]
\centering
\includegraphics[width=\linewidth]{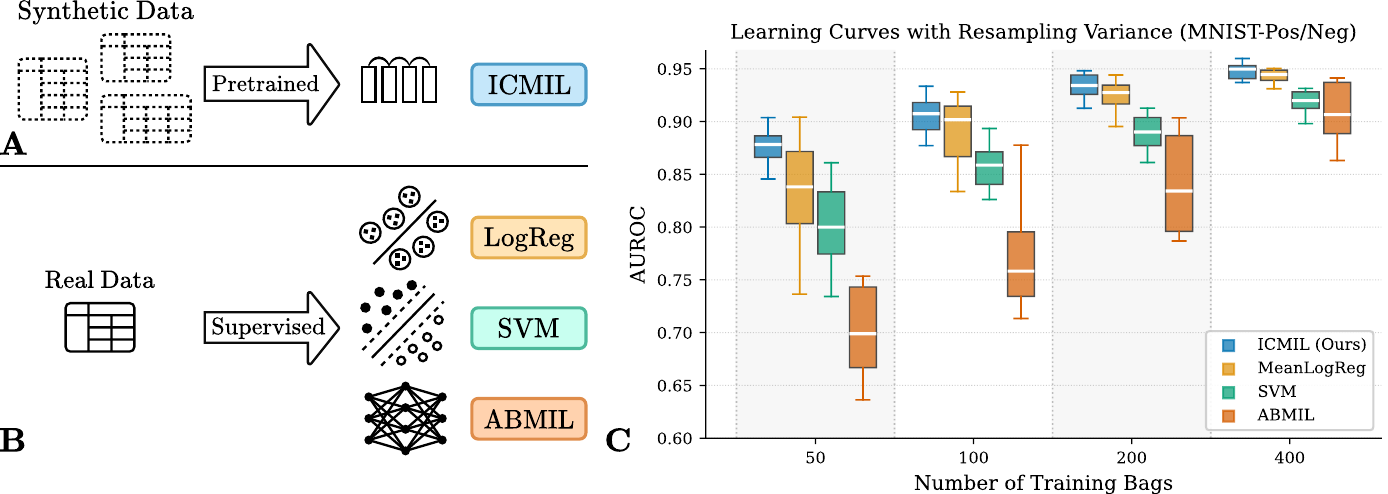}
\caption{(\textbf{A}) ICMIL is pretrained on synthetic bag-structured data and classifies new tasks in a single forward pass without gradient updates or hyperparameter tuning. (\textbf{B}) Supervised methods (MeanLogReg, SVM, ABMIL) are retrained and tuned from scratch on each downstream dataset. (\textbf{C}) Resampling variance on MNIST-Pos/Neg, a binary bag classification task where the label depends on the balance of positive and negative digit classes. Boxplots show the distribution of AUROC over 20 independently resampled training sets with a fixed test set. ICMIL achieves the highest median AUROC and the lowest variance across training set draws, with the advantage most pronounced in the low-bag regime.}
\label{fig1}
\end{figure}

To close this gap, we introduce \emph{In-Context Multiple Instance Learning} (ICMIL). Specifically, we build on the Prior-data Fitted Network (PFN) paradigm~\cite{muller_transformers_2022} and train a model on diverse synthetic bag-structured data to approximate the posterior predictive distribution over bag labels. Intuitively, by training across many simulated MIL tasks, the model learns what plausible bag classifiers look like, so at test time it can infer the labeling rule and use it to predict labels for unseen bags. This shifts the problem from fitting an aggregator on a handful of bags to designing good data generators (priors) over bag-structured tasks that span the relevant hypothesis space. At inference time, the model is given a real MIL dataset as context and classifies the test bags in a single forward pass, without gradient updates or task-specific training. The trained model approximates Bayesian model averaging over hypotheses consistent with the labeled context and avoids the variance that comes with fitting a single aggregator on scarce labels~\cite{geman_neural_1992}. We illustrate this favorable bias-variance trade-off in Figure~\ref{fig1} and show that our ICMIL achieves both higher median AUROC and lower variance across training-set resamples than supervised baselines in the low-sample regime. 

While PFNs have been successfully applied to flat tabular classification \cite{hollmann_tabpfn_2023, qu_tabicl_2025} and regression \cite{hollmann_accurate_2025}, adapting them to bag-structured data requires designing both architectures suited to in-context learning over hierarchical sets and synthetic priors that capture MIL reasoning patterns. On the architectural side, challenges are that (i) attention over all instances across all bags scales quadratically in both, (ii) bag-level compression should be task-aware and take bag labels into account, and (iii) the model should be permutation-invariant within bags while preserving bag identity across them. In this paper, we propose to address these with a Perceiver-style architecture in which learnable bag tokens iteratively gather information from their instances and exchange it with representations and labels of other bags. 

On the prior side, it is unclear what a good data generator for MIL problems looks like. Standard MIL architectures assume that bag labels factorize through a permutation-invariant decomposition over instances, and the natural question is whether this assumption should also be reflected in the prior. We design two families of generators that take opposite stances. Factorized priors encode the classical decomposition, while joint priors drop it in favor of a single generator over the full bag that permits richer dependence structures among instances. We study whether either family suffices on its own or whether a mixture yields a single model that performs robustly across MIL regimes.

We summarize our contributions as follows:
\begin{itemize}
    \item We introduce \emph{In-Context Multiple Instance Learning} (ICMIL) and identify key architectural challenges that arise when dealing with bag-structured data. We propose a Perceiver-style architecture for hierarchical set inputs as a solution.
    \item We propose data generators that encode different MIL assumptions and find that they yield models with complementary strengths. Each model excels on a distinct subset of real benchmarks.
    \item We show that ICMIL, a model trained on a mixture of synthetic priors, achieves the best average AUROC and rank across twelve MIL benchmarks in the low-label regime. It outperforms supervised baselines despite using no gradient updates or hyperparameter tuning.
\end{itemize}

\section{Background}
\label{sec:background}
 
\subsection{Multiple Instance Learning}

In MIL, each data point is a \emph{bag} $B = \{x_1, \ldots, x_I\}$ of $I$ instances $x_i \in \mathbb{R}^F$, annotated with a single bag-level label $y$, which may represent a classification, regression, or time-to-event target. The number of instances $I$ may vary across bags. The original MIL formulation \cite{dietterich_solving_1997, maron_mil_1997} assumes that instances within a bag are statistically independent and permutation-invariant, and that each instance carries an unknown binary label $y_i \in \{0, 1\}$ determining the bag label via $y = \max_i \{y_i\}$. These assumptions have been relaxed to accommodate a wider range of applications, including more general bag-label functions capturing complex instance-label relationships \cite{foulds_review_2010, carbonneau_multiple_2018, hense_xmil_2024}, or instance dependencies arising from shared generative structure such as spatial proximity in pathological tissue~\cite{zhao_predicting_2020, shao_transmil_2021}. Traditionally, MIL models rely on pre-defined pooling functions, such as max or mean pooling, to aggregate instance predictions into a bag prediction \cite{dietterich_solving_1997,campanella_clinical_2019}. \citet{ilse_attention_2018} introduced learned instance aggregation via attention, enabling the model to identify instances most predictive for a given task. A plethora of extensions have been proposed, e.g., exploiting self-attention to explicitly model dependencies between instances~\cite{shao_transmil_2021}. However, these approaches depend on weakly supervised learning from task-specific labeled bags, making them prone to overfitting and shortcut learning \cite{howard_signatures_2021, koemen_towards_2025, drexlin_medi_2025, mollers_mind_2026, dawood_confounding_2026}, particularly when training data is scarce. To reduce reliance on labeled training data, several directions for zero- or few-shot MIL have been explored: leveraging vision-language models \cite{lu_visual_2023, meseguer_mil-adapter_2026}, self-supervised pretraining \cite{xu_whole-slide_2024, ding_multimodal_2025}, and transfer learning across related tasks~\cite{shao_do_2025}. However, such methods constrain the flexibility of the aggregation mechanism and remain dependent on domain-specific pretraining, limiting their applicability across diverse tasks and domains. In contrast, our ICL approach requires no task-specific labeled training data and generalizes across tasks and application domains without finetuning. By combining cross-bag, cross-instance, and cross-feature attention, ICMIL enables both task-specific instance aggregation and effective few-shot inference.

\subsection{Prior-data Fitted Networks}

Prior-data Fitted Networks (PFNs)~\cite{muller_transformers_2022} amortize Bayesian inference over a prior of data-generating processes via in-context learning. A transformer is pretrained on synthetic datasets sampled from a prior over data-generating processes to predict labels of held-out points given the remaining examples as context. At inference time, a PFN directly outputs predictions for a new dataset in a single forward pass, without gradient-based optimization. Performance depends on the alignment between the prior and the target task, motivating careful prior design. Recently, PFNs have matched or surpassed classical methods on tabular prediction benchmarks~\cite{hollmann_tabpfn_2023,hollmann_accurate_2025} and have been extended to causal inference~\citep{robertson_do-pfn_2025}, time series forecasting~\cite{hoo_tables_2026}, Bayesian optimization~\cite{muller_pfns4bo_2023} and single-cell perturbation effect estimation~\cite{sextro_mappfn_2026}. Closest to our work, \citet{kopp2025utilizing} adapts TabPFN to multi-instance regression by collapsing each bag into a tabular input through k-means pooling. However, their aggregation is unsupervised and fixed before inference, which limits the bag-label functions it can represent (Appendix~\ref{app:aggregation}). In contrast, we directly train a PFN over bag-structured inputs with priors and an architecture that can use MIL labels in-context.

\section{In-Context Learning for Multiple Instance Problems}

We now formulate the in-context learning problem for MIL. Let 
$p(\mathcal{D})$ be a prior over MIL datasets that allows us to sample a dataset $\mathcal{D} = \{(B_i, y_i)\}_{i=1}^{N+1}$ of bags and labels. The learner $q_\theta$  receives the full set of $N$ labeled context bags as direct input and has to infer $y_{N+1}$ for a query bag $B_{N+1}$. We train $q_\theta$ to minimize the expected negative log-likelihood
\begin{equation}
    \mathcal{L}(\theta) = \mathbb{E}_{\mathcal{D} \sim p(\mathcal{D})}
    \left[ -\log \, q_\theta(y_{N+1} \mid B_{N+1},
    \{(B_i, y_i)\}_{i=1}^{N}) \right].
\end{equation}
To design a learner that minimizes the objective, one has to take some particular challenges into account that come with the bag-structured nature of MIL data.

\paragraph{Challenge 1: Computational scalability.}

A dataset of $N$ bags with $I$ instances contains $N \times I$ feature vectors. Naive attention over all instances across all bags scales as $\mathcal{O}(N^2 I^2)$ in compute. This quickly becomes prohibitively expensive as MIL datasets frequently have hundreds or thousands of instances per bag.  

\paragraph{Challenge 2: Task-dependent instance compression.}

A naive solution to Challenge 1 would be to first aggregate instance-level features into bag representations, and then perform a full attention over these compressed representations and the corresponding labels. In this case, compression happens before the learner can attend across labels and bags, and cannot use task-specific information to determine which instance-level features to preserve. For example, the same bag of tissue patches could be used for a tumor detection task because of a single atypical cell, or for a subtyping task because of the overall tissue architecture. Any compression layer must therefore be able to incorporate label information across all bags to decide which instance-level features to preserve. 

\paragraph{Challenge 3: Permutation invariance while preserving bag identity.}

Datasets in typical MIL problems are often sets. That is, there is no specific ordering of the instances of a bag and the model should be invariant to any permutation of the instances within bags of a dataset. At the same time, the model needs to recognize which instances belong to the same bag. 

\begin{figure}[h]
    \centering
    \includegraphics[width=\linewidth]{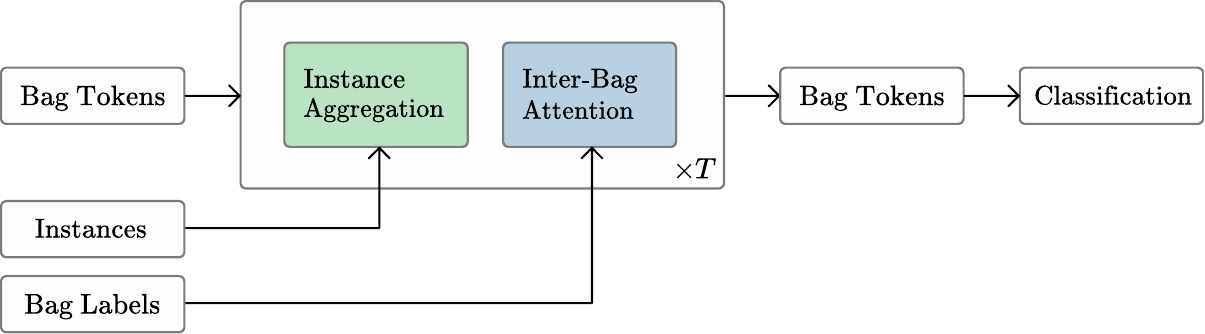}
    \caption{Schematic of the ICMIL architecture, illustrated for $T$ iterations. A learnable bag token $\mathbf{Q}_b \in \mathbb{R}^{G \times E}$ is initialized for each bag. At every iteration, each bag token is first updated by \textbf{instance aggregation}, where it cross-attends to its own instance tokens (applied independently per bag). It is then concatenated with the bag-label embedding $\tilde{\mathbf{y}}_b$ and updated by \textbf{inter-bag attention}, a column-row self attention over all bags. After $T$ iterations, the label token of each query bag is decoded into the predicted class distribution.}
    \label{fig2:architecture}
\end{figure}

\subsection{An Architecture for ICMIL} \label{sec:arch}

We now show how we can design a Perceiver-style~\cite{jaegle_perceiver_2021} architecture that resolves the above challenges. To implement the architecture, we follow \cite{hollmann_accurate_2025} and embed each of the $F$ input features into groups of size $s$, yielding $G = \lceil F / s \rceil$ feature-group embeddings of dimension $E$ per instance. This produces a tensor $\tilde{\mathbf{X}} \in \mathbb{R}^{N \times I \times G \times E}$ of instance embeddings. We further embed the bag labels into $\tilde{\mathbf{y}} \in \mathbb{R}^{N \times E}$, with test bags receiving the mean of the training label embeddings. We then instantiate a learnable token $Q_{b}\in \mathbb{R}^{G \times E}$ for each bag and update it over $T$ steps of alternating cross attention from a bag vector to its instance vectors and self attention among bag vectors (Figure~\ref{fig2:architecture}): 

\paragraph{Instance aggregation / Cross attention: Bag tokens attend to instance tokens.}  The $G$ latent bag-level tokens $\mathbf{Q}_{b,g} \in \mathbb{R}^{E}$ are updated by attending to their respective instance tokens across the corresponding feature group:
\begin{equation}
    {\mathbf{Z}}_{b,g}^{(t)} = \mathbf{Q}_{b,g}^{(t)} + 
    \mathrm{MHA}\!\left(\mathbf{Q}_{b,g}^{(t)},\, 
    \tilde{\mathbf{X}}_{b,g},\, \tilde{\mathbf{X}}_{b,g}\right)
\end{equation}
where $\mathrm{MHA}(Q, K, V)$ denotes multi-head attention and $\tilde{\mathbf{X}}_{b,g} \in \mathbb{R}^{I \times E}$ are the instance tokens of bag $b$ for feature group $g$. Since the operation is independent per bag, we can chunk over the bags $N$ and process only one bag at a time, reducing peak memory from $\mathcal{O}(N \cdot I \cdot G)$ to $\mathcal{O}(I \cdot G)$.

\paragraph{Inter-bag attention / Self attention: Bag tokens attend to each other.}  After instance aggregation, the label embedding $\tilde{\mathbf{y}}_b \in \mathbb{R}^{E}$ is appended to $\mathbf{Z}_{b}^{(t)}$, so that each bag is represented by $G + 1$ tokens of size $E$. Stacked across all bags, this yields a matrix $\hat{\mathbf{Z}}^{(t)} \in \mathbb{R}^{N \times (G+1) \times E}$ that is processed with column-row attention~\cite{hollmann_accurate_2025}:
\begin{equation}
    \mathbf{Q}^{(t+1)} = \mathrm{ColRowAttn}\!\left(\hat{\mathbf{Z}}
    \right)
\end{equation}
where column attention operates over feature groups and labels, and row attention allows for the flow of information between bags. Importantly, test bags can only attend to training context bags, but not to each other. The compute for this step scales as $\mathcal{O}(N^2 + G^2)$, with memory linear in $\mathcal{O}(N \cdot G)$ using flash attention. Removing this step makes in-context learning impossible (Appendix~\ref{app:ablations}).

After $T$ iterations, the label token of each query bag is passed through a decoder to produce the predicted class distribution. The architecture directly addresses the three challenges identified above. Importantly, instance aggregation reduces the cost and complexity compared to naive attention over all instances (Challenge~1). Furthermore, the iterative attention mechanism naturally resolves the task-dependent compression problem. The model does not need to compress instances into bag representations in a single label-agnostic step, but can alternate between attending to instances and attending across bags and labels (Challenge~2). Finally, since $\mathrm{MHA}$ is permutation-invariant over its inputs, the 
architecture is invariant to instance ordering within bags while preserving bag identity across bags (Challenge~3). Appendix~\ref{app:ablations} validates these choices empirically, showing that both the iterative aggregation and the attention between bags are necessary.

\subsection{Synthetic Priors for ICMIL} \label{met:priors}
 
\begin{figure}[t]
  \centering
  \includegraphics[width=\linewidth]{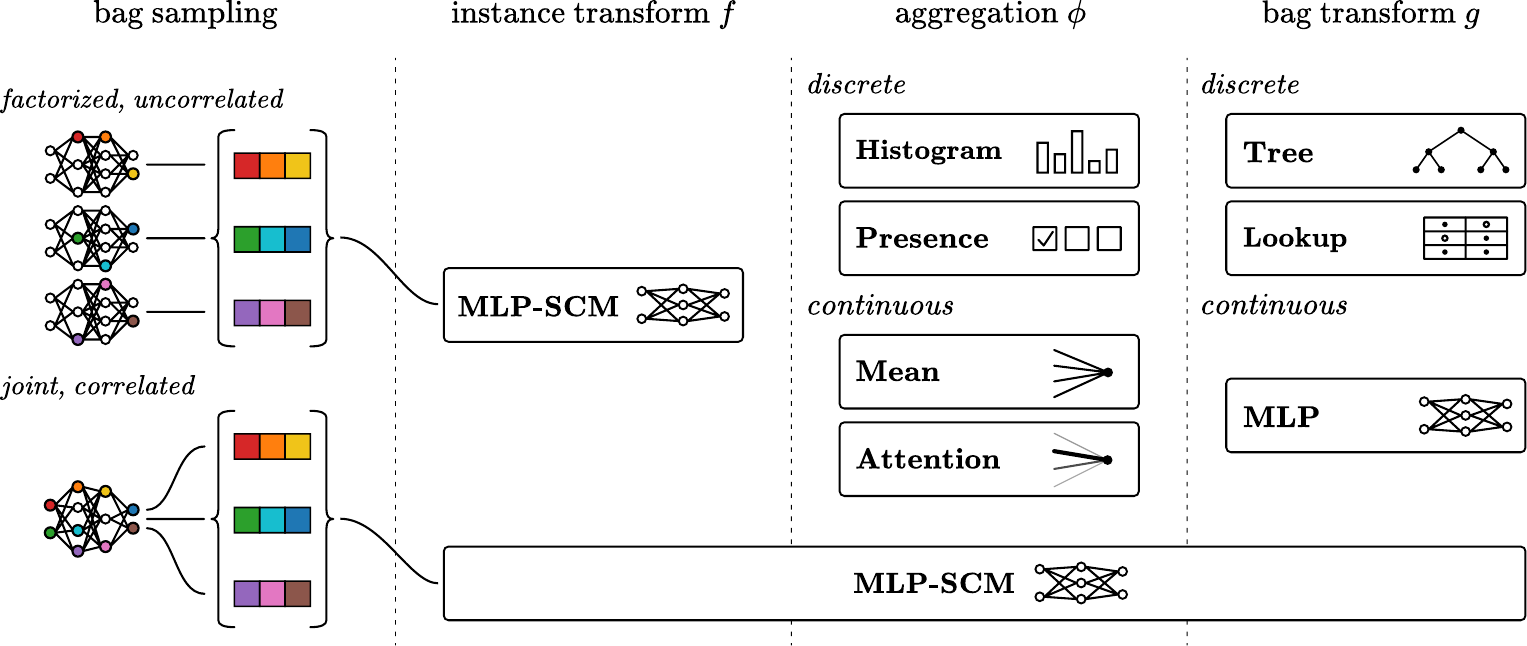}
  \caption{Taxonomy of bag-structured priors for ICMIL. \textbf{Bag sampling}: factorized priors draw each instance from an independent SCM (top), producing uncorrelated bags, while joint priors draw all instances from a single SCM (bottom), inducing within-bag feature correlations. \textbf{Instance transform} $f$: 
  induced by the SCM as the mapping from the feature nodes $x_i$ to a designated output node along the causal graph.
  \textbf{Aggregation} $\phi$: for factorized priors, the per-instance outputs are compressed by a continuous or discrete permutation-invariant summary. \textbf{Bag transform} $g$: the bag summary is decoded into a label via a tree, a lookup table over discrete class patterns, or an MLP.}
  \label{fig:prior_taxonomy}
\end{figure}

The effectiveness of ICMIL depends on the alignment between its training prior $p(\mathcal{D})$ and the structure of real MIL tasks. We therefore investigate the effect of various prior designs that capture different distributional properties of bag-structured data. We broadly divide our priors into factorized and joint priors, illustrated in Figure~\ref{fig:prior_taxonomy}.

\paragraph{Factorized priors.} These priors follow the dominant MIL modeling approach and assume the bag label is determined by a permutation-invariant function of the
instances~\cite{zaheer_deep_2017}:
\begin{equation}
\label{eq:mil_fg}
    y = g\!\left(\phi\!\left(\{f(x_i)\}_{i=1}^{I}\right)\right),
\end{equation}
where $f\colon \mathbb{R}^F \to \mathbb{R}^d$ is an instance-level transform,
$\phi$ is an aggregation that compresses the instance outputs into a bag summary, and $g$ maps that summary to the bag label.  For each instance, we use an MLP-SCM~\cite{qu_tabicl_2025} with randomly parameterized nonlinear relationships to jointly generate both the instance features $x_i \in \mathbb{R}^F$ and the instance-level outputs $f(x_i) \in \mathbb{R}^d$ from the same causal graph. Latent causes are sampled independently per instance, so that $p(x_1, \ldots, x_I \mid y) = \prod_i p(x_i \mid y)$. For $\phi$ we consider \emph{discrete} and \emph{continuous} summaries. Discrete summaries map the output of $f$ to one of $K$ discrete classes and compute either a histogram of class counts or a presence indicator of which classes appear. Continuous summaries pool the raw instance embeddings via mean pooling or ABMIL pooling~\cite{ilse_attention_2018}. For $g$ we investigate an MLP, a tree, and a lookup table over discrete instance-class patterns.

\paragraph{Joint priors.}
These priors drop the $f$-$g$ decomposition and instead define a single function over the flattened bag $[x_1; \ldots; x_I] \in \mathbb{R}^{I \cdot F}$:
\begin{equation}
\label{eq:mil_joint}
    y = f_{\mathcal{S}}(x_1, \ldots, x_I).
\end{equation}
We use an MLP-SCM~\cite{qu_tabicl_2025} to jointly generate the concatenated instance features $[x_1; \ldots; x_I]$ and the bag label $y$ from the same causal graph, with latent causes sampled once per bag. This allows for inter-instance correlations that the factorized prior cannot express, and in general we have $p(x_1, \ldots, x_I \mid y) \neq \prod_i p(x_i \mid y)$.

The two prior families differ in important ways that may influence downstream performance. Factorized priors are structurally aligned with classical MIL assumptions (permutation invariance, instance-level decomposition, witness-style aggregation), while joint priors are more expressive and induce within-bag feature correlations that the factorized family cannot express. We empirically investigate whether structural alignment or expressiveness leads to better downstream performance, and whether the resulting differences in feature correlation structure influence which prior is best suited to a given MIL task.

\section{Experiments}

We evaluate the proposed in-context learner on twelve MIL benchmarks. We investigate (i) whether the different priors encode complementary inductive biases and MIL regimes, (ii) whether a single model trained on a mixture of priors can inherit their individual strengths, and (iii) whether that model is more sample-efficient than standard MIL baselines in the low-label regime that frequently occurs in practical applications. We make our code available at \url{https://github.com/injurise/ICMIL}.

\subsection{Experimental Setup}

\paragraph{Training.}
We pre-generate synthetic datasets from each prior configuration and store them separately. For both the MLP and Tree components of our priors, we adapt the implementation from \citet{qu_tabicl_2025} and use their default hyperparameter sampling ranges. The generated data is split into context and test bags and retrieved during training. For the prior ablations, we use a reduced setup with models of 6 iterations, 4 attention heads, embedding dimension 128, MLP hidden size 512, and feature group size 1, trained on instances with 25 features. Models are trained for $20{,}000$ steps with batch size 128. The number of bags per dataset is sampled uniformly in $[70, 125]$. Based on the results in this reduced setting, we train a final model with increased embedding dimension and training duration in \S\ref{sec:exp:lowdata}. Further training details can be found in Appendix~\ref{app:training_details}.

\paragraph{Benchmarks.}
We evaluate on twelve multiple-instance learning benchmarks. Six benchmarks expose a witness label rule (SMIL, Musk1, Musk2, Letters, HEPMASS, RSNA-ICH) and six rely on interactions across instances (Elephant, Fox, Tiger, TCGA, Adjacent Pairs, Pos/Neg).  We subsample large benchmarks (e.g.\ RSNA-ICH, TCGA LUAD-LUSC) and use them to create tasks with approximately 100 bags that fall into the low-sample regime. We apply PCA to the benchmarks whose feature dimension exceeds 25. The full per-benchmark descriptions are located in Appendix~\ref{app:benchmarks}. 

\paragraph{Baselines.}
We compare our method against eight deliberately heterogeneous MIL baselines.
\textbf{MeanLogReg:} We fit scikit-learn's \texttt{LogisticRegressionCV} on the mean of the features of each bag. We apply $5$-fold stratified CV over $C\in\{10^{-2},10^{-1},1,10,10^{2}\}$.
\textbf{SVM-Summ:} For each feature we compute six fixed summary statistics (sum, mean, median, min, max, stdv) and concatenate them to form one vector per bag. We then fit an RBF-kernel SVM, with $C\in\{10^{-2},10^{-1},1,10,10^{2}\}$ tuned by stratified $5$-fold cross-validation.
\textbf{ABMIL:} We train an ABMIL model with embedding dimension $500$ and attention dimension $128$, selecting the learning rate from ${10^{-2},5{\times}10^{-3},10^{-3},5{\times}10^{-4},10^{-4}}$ and weight decay from ${0,10^{-4},5{\times}10^{-4}}$ using a single stratified $10\%$ held-out validation split. Each candidate is trained for up to $200$ epochs in full-batch mode with Adam and early stopping (patience $20$). The configuration with the lowest validation cross-entropy is used as the final model.
\textbf{DSMIL:} We train the dual-stream model of \citet{li_dual-stream_2021} with embedding dimension $512$ and attention dimension $384$, and tune the learning rate and weight decay with the same protocol as for ABMIL.
\textbf{ACMIL:} We train the gated-attention model of \citet{zhang_attention_2024} with embedding dimension $256$, attention dimension $128$, $5$ attention branches, top-$10$ masked patches and a mask-drop rate of $0.6$, and tune the learning rate and weight decay with the same protocol.
\textbf{TabPFN-Concat:} We flatten each bag to a single vector by concatenating the instances and pass it through TabPFN-v2 \citep{hollmann_accurate_2025}. In cases where the flattened vector exceeds $500$ features we truncate to avoid OOM errors.
\textbf{TabPFN-Subsample:} We randomly subsample whole instances per bag, flatten them and predict a label. We repeat this $10$ times and average the logits to obtain the final prediction.
\textbf{TabPFN-Cluster:} Following \citet{kopp2025utilizing}, we cluster all training instances with $K$-means ($K{=}5$). We then assign all instances in the bags a cluster label. For each bag, the instances belonging to the same cluster are summed and scaled by the ratio of total bag size to cluster instance count. This produces one feature vector per bag per cluster. TabPFN-v2 is then called for each cluster vector and bag giving logits for each cluster. The final prediction is the average of the resulting logits across all clusters the bag participates in.

\begin{figure}
  \centering
  \begin{minipage}[c]{0.55\linewidth}
  \centering
  \setlength{\tabcolsep}{3pt}
  \resizebox{\linewidth}{!}{%
  \begin{tabular}{lccccc}
  \toprule
  Prior & Wit/Uncorr & Wit/Corr & Int/Uncorr & Int/Corr & Overall \\
  \midrule
  Joint & $2.60$ & $\fakebf{1.00}$ & $\fakebf{1.00}$ & $\fakebf{1.75}$ & $\fakebf{1.92}$ \\
  Fact. $(\mathrm{cont},\mathrm{MLP})$ & $\fakebf{1.80}$ & $\underline{2.00}$ & $\underline{2.25}$ & $\underline{2.12}$ & $\underline{2.00}$ \\
  Fact. $(\mathrm{cont},\mathrm{tree})$ & $3.40$ & $3.00$ & $2.75$ & $2.75$ & $3.04$ \\
  Fact. $(\mathrm{disc},\mathrm{lookup})$ & $\underline{2.40}$ & $4.00$ & $4.00$ & $3.38$ & $3.12$ \\
  Fact. $(\mathrm{disc},\mathrm{MLP})$ & $4.80$ & $5.00$ & $5.00$ & $5.00$ & $4.92$ \\
  Fact. $(\mathrm{disc},\mathrm{tree})$ & $6.00$ & $6.00$ & $6.00$ & $6.00$ & $6.00$ \\
  \bottomrule
  \end{tabular}%
  }
  \end{minipage}%
  \hfill
  \begin{minipage}[h]{0.43\linewidth}
  \centering
  \includegraphics[width=\linewidth]{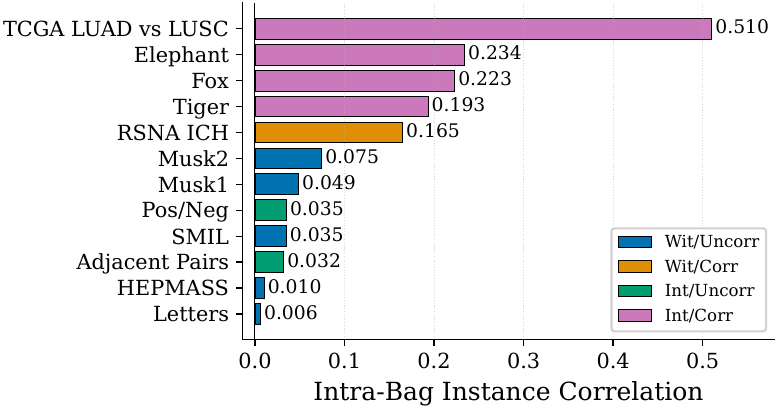}
  \end{minipage}
  \caption{\textbf{Left:} Mean rank across MIL benchmarks for models trained on joint and factorized priors, broken down by group and pooled over all datasets (Overall). Lower is better. Wit/Corr contains a single benchmark (RSNA-ICH). \textbf{Right:} Within-bag feature correlation (ICC) per benchmark and colored by task groups. ICC quantifies the share of feature variance explained by bag membership. High values indicate strong within-bag correlation.}
  \label{fig:prior_ranks_with_png}
\end{figure}

\subsection{Different Priors Capture Different MIL Regimes} \label{sec:exp:priors}

We train one model per prior configuration with the reduced setup and report per-benchmark AUROC scores in Table~\ref{tab:prior_results}. We group the benchmarks by the label rule (witness vs.\ interaction) and the within-bag feature correlation, measured by the intra-class correlation coefficient (ICC) of instance features (Figure~\ref{fig:prior_ranks_with_png}). 

The \textsc{Joint} prior achieves the best average rank and is especially strong in settings where within-bag correlation and/or interaction are present. Nevertheless, the factorized priors outperform it by large margins on the uncorrelated witness benchmarks. The \textsc{Factorized}$(\text{disc},\text{lookup})$ prior that was explicitly designed for such tasks has a large advantage on Letters ($95.2 \pm 0.5$), SMIL ($74.9 \pm 0.3$) and HEPMASS ($86.1 \pm 1.0$) where it improves over the \textsc{Joint} prior by up to $4$ percentage points. The \textsc{Factorized}$(\text{continuous},\text{MLP})$ prior takes the lead on Musk1 ($93.9 \pm 0.6$) and Musk2 ($85.4 \pm 3.2$). The \textsc{Factorized}$(\text{disc},\text{tree})$ prior collapses to near-chance performance across all benchmarks, indicating that this combination is not suitable for real MIL tasks.

These results address the questions raised in \S\ref{met:priors}. The more expressive joint prior achieves the best average rank, but the factorized priors that are structurally aligned with the MIL paradigm outperform it on uncorrelated witness benchmarks. Interestingly, the factorized priors do not have an advantage on the witness task RSNA-ICH where features are correlated. This suggests that the correlation structure of the instance features might play a larger role than the label rule in determining the optimal prior.
Overall, we observe that greater expressiveness does not always translate into better downstream performance and that aligning priors to target tasks can yield measurable advantages. The fact that no single prior dominates motivates the prior-mixing experiment in the following section.

\begin{table}[t]
  \centering
  \setlength{\tabcolsep}{3pt}
  \renewcommand{\arraystretch}{1.1}
  \setlength{\aboverulesep}{0pt}\setlength{\belowrulesep}{0pt}
  \resizebox{\textwidth}{!}{%
  \begin{tabular}{l | cc | cccccccccccc}
  \toprule
  \headstrut Model & \rotatebox{90}{Avg AUROC} & \rotatebox{90}{Avg Rank} & \rotatebox{90}{Musk1} & \rotatebox{90}{Musk2} & \rotatebox{90}{HEPMASS} & \rotatebox{90}{Letters} & \rotatebox{90}{SMIL} & \rotatebox{90}{Adj.\ Pairs} & \rotatebox{90}{Pos/Neg} & \rotatebox{90}{RSNA ICH} & \rotatebox{90}{Elephant} & \rotatebox{90}{Fox} & \rotatebox{90}{Tiger} & \rotatebox{90}{TCGA} \\
  \midrule
    \multicolumn{15}{>{\columncolor{gray!20}}l}{\rowstrutc\textit{Baselines}} \\
    \rowstrutc MeanLogReg & \sv{82.37} & \sv{7.46} & \val{91.6}{0.5} & \val{82.1}{1.6} & \valu{86.9}{0.3} & \val{92.3}{1.0} & \val{72.4}{0.2} & \val{73.9}{0.1} & \val{85.4}{0.5} & \val{74.6}{0.0} & \val{93.3}{0.1} & \val{62.5}{0.6} & \val{86.3}{0.2} & \val{87.1}{0.2} \\
    \rowstrutc SVM-Summ & \sv{79.38} & \sv{7.08} & \valb{94.7}{0.2} & \val{81.3}{0.7} & \val{42.1}{5.1} & \val{93.9}{0.5} & \val{66.5}{1.9} & \val{77.7}{0.8} & \val{84.5}{0.2} & \valb{77.0}{0.4} & \val{92.5}{0.1} & \valu{65.7}{0.7} & \val{88.5}{0.7} & \val{88.2}{0.2} \\
    \rowstrutc ABMIL & \sv{79.97} & \sv{9.17} & \val{85.7}{3.5} & \val{75.4}{2.8} & \val{75.7}{1.7} & \val{97.6}{0.8} & \valu{85.4}{0.5} & \val{65.5}{1.1} & \val{74.6}{1.6} & \val{71.2}{0.3} & \val{94.1}{0.2} & \val{59.8}{1.6} & \val{83.9}{1.5} & \valb{90.7}{1.6} \\
    \rowstrutc DSMIL & \sv{80.33} & \sv{8.58} & \val{86.2}{1.8} & \val{84.0}{3.9} & \val{84.5}{1.1} & \val{97.6}{1.2} & \val{80.0}{1.7} & \val{65.2}{1.4} & \val{69.9}{1.5} & \val{69.5}{0.7} & \val{93.4}{0.4} & \val{61.2}{0.6} & \val{84.7}{3.0} & \val{87.8}{1.9} \\
    \rowstrutc ACMIL & \svu{83.81} & \sv{5.88} & \val{89.0}{1.3} & \val{83.9}{2.5} & \val{81.1}{3.9} & \valu{98.7}{1.3} & \valb{87.7}{1.0} & \val{72.5}{0.6} & \val{80.8}{0.4} & \val{74.8}{1.0} & \val{94.1}{0.2} & \val{64.4}{1.7} & \val{88.0}{1.2} & \valb{90.7}{0.7} \\
    \rowstrutc TabPFN-Concat & \sv{74.16} & \sv{11.50} & \val{91.1}{0.0} & \val{79.3}{0.0} & \val{66.4}{0.0} & \val{88.8}{0.0} & \val{56.3}{0.0} & \val{57.5}{0.0} & \val{63.7}{0.0} & \val{63.9}{0.0} & \val{91.4}{0.0} & \val{63.9}{0.0} & \valb{89.6}{0.0} & \val{78.0}{0.0} \\
    \rowstrutc TabPFN-Subsample & \sv{74.82} & \sv{10.67} & \val{92.5}{0.2} & \val{78.6}{6.2} & \val{67.7}{0.3} & \val{88.8}{0.0} & \val{58.2}{0.2} & \val{57.0}{0.1} & \val{65.4}{0.3} & \val{64.1}{0.0} & \val{92.0}{0.0} & \valu{65.7}{0.2} & \valb{89.6}{0.1} & \val{78.2}{0.1} \\
    \rowstrutc TabPFN-Cluster & \sv{81.21} & \sv{6.67} & \val{88.0}{1.0} & \valb{91.3}{0.4} & \val{67.5}{3.0} & \valb{99.2}{0.8} & \val{77.8}{1.0} & \val{71.4}{0.4} & \val{77.0}{1.8} & \valu{76.9}{0.1} & \valb{95.6}{0.2} & \valb{66.2}{0.0} & \val{86.5}{0.3} & \val{77.1}{0.2} \\
  \midrule
    \multicolumn{15}{>{\columncolor{gray!20}}l}{\rowstrutc\textit{Prior Ablations (reduced setup)}} \\
    \rowstrutc Joint & \sv{82.56} & \sv{5.79} & \val{91.0}{0.3} & \val{83.2}{3.4} & \val{81.9}{1.1} & \val{90.9}{1.0} & \val{74.5}{0.6} & \valb{78.3}{0.6} & \val{86.9}{0.5} & \val{74.9}{0.3} & \valu{95.0}{0.4} & \val{56.1}{3.4} & \valu{89.2}{0.1} & \valu{88.8}{0.2} \\
    \rowstrutc Fact.~$(\mathrm{cont},\mathrm{MLP})$ & \sv{82.76} & \sv{6.50} & \valu{93.9}{0.6} & \val{85.4}{3.2} & \val{84.3}{0.7} & \val{92.3}{1.0} & \val{71.4}{0.8} & \valu{77.9}{0.1} & \val{86.3}{0.3} & \val{73.9}{0.2} & \val{93.7}{0.3} & \val{58.2}{2.0} & \val{88.9}{0.1} & \val{86.9}{0.4} \\
    \rowstrutc Fact.~$(\mathrm{cont},\mathrm{tree})$ & \sv{81.55} & \sv{8.71} & \val{90.3}{1.0} & \val{82.5}{1.5} & \val{84.0}{0.5} & \val{87.5}{0.3} & \val{70.3}{0.5} & \valu{77.9}{0.4} & \val{85.1}{0.4} & \val{73.4}{0.0} & \val{93.1}{0.3} & \val{61.0}{0.4} & \val{87.3}{0.2} & \val{86.2}{0.8} \\
    \rowstrutc Fact.~$(\mathrm{disc},\mathrm{lookup})$ & \sv{80.98} & \sv{9.17} & \val{87.0}{2.4} & \val{75.2}{0.8} & \val{86.1}{1.0} & \val{95.2}{0.5} & \val{74.9}{0.3} & \val{74.5}{0.7} & \val{84.9}{0.3} & \val{72.4}{0.2} & \val{92.5}{0.1} & \val{57.3}{0.5} & \val{84.9}{0.8} & \val{86.9}{0.6} \\
    \rowstrutc Fact.~$(\mathrm{disc},\mathrm{MLP})$ & \sv{75.12} & \sv{13.04} & \val{76.7}{4.4} & \val{75.3}{5.2} & \val{80.5}{5.5} & \val{77.9}{6.5} & \val{65.5}{6.6} & \val{71.1}{3.9} & \val{78.7}{6.9} & \val{70.1}{3.5} & \val{87.9}{4.4} & \val{51.6}{2.4} & \val{83.9}{2.3} & \val{82.3}{4.9} \\
    \rowstrutc Fact.~$(\mathrm{disc},\mathrm{tree})$ & \sv{51.27} & \sv{15.92} & \val{40.6}{2.9} & \val{50.4}{3.8} & \val{57.9}{4.9} & \val{54.7}{2.5} & \val{49.0}{0.8} & \val{40.9}{6.6} & \val{51.9}{3.6} & \val{51.3}{0.7} & \val{56.3}{1.4} & \val{49.2}{1.4} & \val{67.3}{5.1} & \val{45.8}{13.2} \\
    \rowstrutc Mixed & \sv{83.38} & \svu{5.21} & \val{92.5}{0.5} & \val{87.6}{1.0} & \valb{87.2}{0.5} & \val{94.4}{1.6} & \val{74.5}{0.5} & \val{77.3}{0.3} & \valb{87.3}{0.2} & \val{74.8}{0.0} & \val{94.5}{0.1} & \val{53.9}{2.1} & \val{88.2}{0.6} & \val{88.3}{0.2} \\
  \midrule
    \multicolumn{15}{>{\columncolor{gray!20}}l}{\rowstrutc\textit{Mixed Prior (scaled setup)}} \\
    \rowstrutc \textbf{ICMIL (Ours)} & \svb{84.17} & \svb{4.67} & \val{93.3}{0.6} & \valu{90.8}{1.1} & \val{84.5}{0.7} & \val{95.7}{0.7} & \val{74.4}{0.3} & \val{77.6}{0.5} & \valu{87.0}{0.2} & \val{75.0}{0.1} & \val{94.2}{0.1} & \val{60.7}{0.8} & \val{88.9}{0.6} & \val{87.9}{0.1} \\
  \bottomrule
  \end{tabular}%
  }
  \caption{Per-benchmark mean AUROC (\%) $\pm$ standard error for each model. \textbf{Bold} marks the best result per benchmark and \underline{underline} the second-best (ties marked for both). The two leftmost numeric columns report each model's mean AUROC across all benchmarks and its mean rank (lower is better). For the prior ablations, we use a reduced setup with $20{,}000$ training steps, embedding dimension $128$, and MLP hidden size $512$, while the scaled setup used for the final ICMIL model doubles each of these.}
  \label{tab:prior_results}
\end{table}

\subsection{Mixing Priors Yields a Robust Generalist} \label{sec:exp:mixing_priors}

The previous results show that priors tailor to specific downstream tasks. We test whether a single model can inherit these per-task strengths by training on a weighted mixture of the three priors with complementary coverage from \S\ref{sec:exp:priors}.  We include \textsc{Joint} with a weight of $0.70$, \textsc{Factorized}$(\text{cont},\text{MLP})$ with $0.15$, and \textsc{Factorized}$(\text{disc},\text{lookup})$ with $0.15$. We select these weights a priori and report results for alternative weightings in Appendix~\ref{app:ablations}. In this way, we hope to preserve the joint prior's strong overall performance while improving on the uncorrelated witness tasks where performance seems to be largely driven by the factorized priors. We report this model as \textsc{Mixed} in Table~\ref{tab:prior_results}. \textsc{Mixed} retains the joint prior's lead on correlated and interaction-driven benchmarks (e.g.\ Pos/Neg $87.3 \pm 0.2$, TCGA $88.3 \pm 0.2$) while recovering most of the gap to the factorized priors on the uncorrelated witness tasks (Letters $94.4 \pm 1.6$ vs.\ $95.2 \pm 0.5$; HEPMASS $87.2 \pm 0.5$, the best result overall). It achieves better overall performance across all benchmarks than any of the individual priors.

\subsection{Scaling Yields Further Improvement on Selected Benchmarks} \label{sec:exp:lowdata}

Building on the \textsc{Mixed} model from \S\ref{sec:exp:mixing_priors}, we ask whether scaling along training duration and capacity yields a pretrained model that outperforms supervised baselines on average in the small-data regime. Specifically, we double the number of training steps to $40{,}000$ at batch size $128$, and increase the embedding dimension to $256$ and the MLP hidden size to $1054$. We refer to the resulting scaled model as ICMIL.

Scaling our setup further improves performance on a subset of the benchmarks. Across the twelve benchmarks, ICMIL achieves the best average AUROC (84.17) and the best average rank (4.67) of any of the models considered in Table~\ref{tab:prior_results}. The strongest supervised baseline is ACMIL ($83.81$), which ICMIL leads by $0.36$ points while requiring no gradient updates or per-dataset tuning. We observe that the improvement from scaling over the \textsc{Mixed} model is driven by Fox ($+6.8$), Musk2 ($+3.2$) and, to a smaller degree, Letters ($+1.3$) and Musk1 ($+0.8$) ~(Figure ~\ref{fig:icmil_benchmarks}). Performance on the remaining benchmarks stays within standard error of \textsc{Mixed}, with the exception of HEPMASS, where we observe a $2.7$-point regression.

\begin{wrapfigure}{r}{0.55\textwidth}
    \centering
    \includegraphics[width=0.50\textwidth]{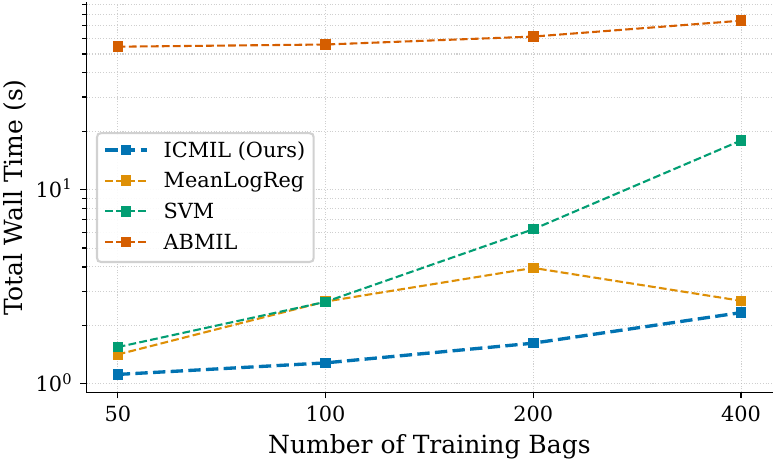}
    \vspace{-10pt}
    \caption{Total wall-clock time on the Pos/Neg benchmark for ICMIL and three supervised MIL baselines. ABMIL is substantially slower due to its per-split hyperparameter search. Once trained, ICMIL classifies all query bags in a single forward pass and is competitive with the simpler linear and kernel baselines.}
    \label{fig:wall_clock_time}
    \vspace{-8pt}
\end{wrapfigure}

Comparing ICMIL to the baselines, we find that no single baseline is competitive across all benchmarks. The strongest aggregate baseline is MeanLogReg with an average AUROC of $82.37$, followed by TabPFN-Cluster ($81.21$) and SVM-Summ ($79.38$). The two flat TabPFN baselines underperform substantially, suggesting that flattening MIL inputs into tabular form discards information that bag-aware architectures can exploit. 
ABMIL produces strong results on TCGA ($90.7 \pm 1.6$) and SMIL ($85.4 \pm 0.5$), and is competitive on Letters ($97.6 \pm 0.8$) and Elephant ($94.1 \pm 0.2$). However, it underperforms on benchmarks like Adjacent Pairs ($65.5 \pm 1.1$) and Musk2 ($75.4 \pm 2.8$). ACMIL, which is explicitly designed to counteract attention overfitting, is markedly more robust in this regime ($83.81$ on average), while DSMIL ($80.33$) stays close to ABMIL and falls off on Adjacent Pairs ($65.2 \pm 1.4$) and Pos/Neg ($69.9 \pm 1.5$). This illustrates the instability caused by the small number of training bags in our benchmarks (typically around $100$), which makes hyperparameter selection and early stopping highly unreliable. ICMIL avoids this failure mode because it requires no hyperparameter selection or task-specific training. This also has practical runtime consequences, as the per-split hyperparameter search of supervised baselines such as ABMIL dominates total wall-clock time, while ICMIL's runtime is competitive with the linear and kernel baselines (Figure~\ref{fig:wall_clock_time}). That said, ICMIL does not dominate on every benchmark. On SMIL, TCGA, and Letters, the baselines still hold a notable lead (e.g.\@ ACMIL on SMIL by $+13.3$ and TabPFN-Cluster on Letters by $+3.5$). This suggests room for improvement, for instance through richer priors, longer pretraining, or finetuning.

\section{Limitations}

A few aspects of our setup leave room for further investigation. Finetuning ICMIL on the training partitions of the benchmark datasets to unlock further performance gains remains a promising direction. Furthermore, our benchmarks reduce the feature dimension to 25 via PCA, leaving higher-dimensional foundation-model embeddings as future work. This gap should be addressable by training on higher-dimensional synthetic data and scaling model capacity and training duration \citep{kolberg_tabpfn-wide_2025}. At the same time, our scaling experiment suggests that gains from added capacity and training time are not
fully uniform across benchmarks and hint at interactions between prior mixture, model size and training duration that warrant closer study.

\section{Conclusion}

We introduced \emph{In-Context Multiple Instance Learning} (ICMIL), a framework for solving MIL problems via in-context learning over bag-structured data. To support this, we proposed an architecture that addresses the scalability, task-dependent compression, and permutation-invariance challenges that arise with hierarchical set inputs, and designed a family of synthetic priors that encode complementary inductive biases over MIL tasks. A model trained on a mixture of these priors achieves the best average AUROC and the best average rank across twelve MIL benchmarks in the low-label regime, ahead of both classical and recent supervised MIL methods.

Beyond the scaling directions outlined in the previous section, several orthogonal avenues are likely to yield further gains. On the prior side, our results show that the choice of synthetic data generator matters considerably and that we can fruitfully combine priors that encode complementary inductive biases. Therefore, designing priors aligned with specific application domains \citep{sextro_mappfn_2026}, such as spatial priors for computational pathology or sequence-aware priors for time-resolved tasks \citep{hoo_tables_2026}, is another promising direction. 

Furthermore, inspired by recent work on tabular PFNs~\cite{hollmann_accurate_2025, qu_tabicl_2025}, post-training on real-world MIL corpora could further close the synthetic-to-real gap \citep{garg_real-tabpfn_2025}, and richer architectural variants remain to be explored. We hope ICMIL serves as a starting point for bringing the benefits
of in-context learning to bag-structured prediction problems, and more broadly to settings where labels are scarce and supervision is weak.

\newpage
\section*{Acknowledgements}
The results shown here are in part based upon data generated by the TCGA Research Network: \url{https://www.cancer.gov/tcga}.

\section*{Author Contributions}
\textbf{Alexander M\"ollers:} Conceptualization, Methodology, Software, Validation, Investigation, Formal Analysis, Visualization, Project Administration, Writing -- original draft, Writing -- review \& editing.
\textbf{Marvin Sextro:} Investigation, Software, Validation, Visualization, Writing -- review \& editing.
\textbf{Julius Hense:} Supervision, Writing -- review \& editing.
\textbf{Gabriel Dernbach:} Conceptualization, Methodology, Supervision, Writing -- review \& editing.
\textbf{Klaus-Robert M\"uller:} Supervision, Funding Acquisition, Resources.

\bibliography{references}

@inproceedings{jaegle_perceiver_2021,
  title     = {Perceiver: General Perception with Iterative Attention},
  author    = {Jaegle, Andrew and Gimeno, Felix and Brock, Andy and Vinyals, Oriol and Zisserman, Andrew and Carreira, Joao},
  booktitle = {Proceedings of the 38th {International} {Conference} on {Machine} {Learning}},
  pages     = {4651--4664},
  year      = {2021},
  volume    = {139},
  series    = {Proceedings of Machine Learning Research},
}

@inproceedings{qu_tabicl_2025,
  title     = {{T}ab{ICL}: A Tabular Foundation Model for In-Context Learning on Large Data},
  author    = {Qu, Jingang and Holzm\"{u}ller, David and Varoquaux, Ga\"{e}l and Le Morvan, Marine},
  booktitle = {Proceedings of the 42nd International Conference on Machine Learning},
  pages     = {50817--50847},
  year      = {2025},
  volume    = {267},
  series    = {Proceedings of Machine Learning Research},
}

@inproceedings{hollmann_tabpfn_2023,
  title     = {Tab{PFN}: A Transformer That Solves Small Tabular Classification Problems in a Second},
  author    = {Noah Hollmann and Samuel M{\"u}ller and Katharina Eggensperger and Frank Hutter},
  booktitle = {The Eleventh International Conference on Learning Representations },
  year      = {2023}
}

@inproceedings{muller_transformers_2022,
  title     = {Transformers Can Do Bayesian Inference},
  author    = {Samuel M{\"u}ller and Noah Hollmann and Sebastian Pineda Arango and Josif Grabocka and Frank Hutter},
  booktitle = {International Conference on Learning Representations},
  year      = {2022},
  pages     = {81861--81875}
}

@article{hollmann_accurate_2025,
  title   = {Accurate predictions on small data with a tabular foundation model},
  volume  = {637},
  number  = {8045},
  journal = {Nature},
  author  = {Hollmann, Noah and Müller, Samuel and Purucker, Lennart and Krishnakumar, Arjun and Körfer, Max and Hoo, Shi Bin and Schirrmeister, Robin Tibor and Hutter, Frank},
  year    = {2025},
  pages   = {319--326}
}

@inproceedings{robertson_do-pfn_2025,
	title = {Do-{PFN}: {In}-{Context} {Learning} for {Causal} {Effect} {Estimation}},
	volume = {38},
	booktitle = {Advances in {Neural} {Information} {Processing} {Systems}},
	author = {Robertson, Jake and Reuter, Arik and Guo, Siyuan and Hollmann, Noah and Hutter, Frank and Schölkopf, Bernhard},
	year = {2025},
	pages = {174811--174848},
}

@inproceedings{muller_pfns4bo_2023,
  title     = {{PFN}s4{BO}: In-Context Learning for {B}ayesian Optimization},
  author    = {M\"{u}ller, Samuel and Feurer, Matthias and Hollmann, Noah and Hutter, Frank},
  booktitle = {Proceedings of the 40th International Conference on Machine Learning},
  pages     = {25444--25470},
  year      = {2023},
  volume    = {202},
  series    = {Proceedings of Machine Learning Research}
}

@inproceedings{sextro_mappfn_2026,
  title     = {{MapPFN}: Learning Causal Perturbation Maps in Context},
  author    = {Sextro, Marvin and K\l{}os, Weronika and Dernbach, Gabriel},
  booktitle = {ICLR 2026 Workshop on Generative AI in Genomics ({Gen}\textsuperscript{2})},
  year      = {2026},
}

@inproceedings{shao_do_2025,
  title     = {Do Multiple Instance Learning Models Transfer?},
  author    = {Shao, Daniel and Chen, Richard J. and Song, Andrew H. and Runevic, Joel and Lu, Ming Y. and Ding, Tong and Mahmood, Faisal},
  booktitle = {Proceedings of the 42nd International Conference on Machine Learning},
  pages     = {54219--54238},
  year      = {2025},
  volume    = {267},
  series    = {Proceedings of Machine Learning Research},
}

@article{xu_whole-slide_2024,
  title   = {A whole-slide foundation model for digital pathology from real-world data},
  volume  = {630},
  number  = {8015},
  journal = {Nature},
  author  = {Xu, Hanwen and Usuyama, Naoto and Bagga, Jaspreet and Zhang, Sheng and Rao, Rajesh and Naumann, Tristan and Wong, Cliff and Gero, Zelalem and González, Javier and Gu, Yu and Xu, Yanbo and Wei, Mu and Wang, Wenhui and Ma, Shuming and Wei, Furu and Yang, Jianwei and Li, Chunyuan and Gao, Jianfeng and Rosemon, Jaylen and Bower, Tucker and Lee, Soohee and Weerasinghe, Roshanthi and Wright, Bill J. and Robicsek, Ari and Piening, Brian and Bifulco, Carlo and Wang, Sheng and Poon, Hoifung},
  year    = {2024},
  pages   = {181--188}
}

@article{ding_multimodal_2025,
  title   = {A multimodal whole-slide foundation model for pathology},
  volume  = {31},
  number  = {11},
  journal = {Nature Medicine},
  author  = {Ding, Tong and Wagner, Sophia J. and Song, Andrew H. and Chen, Richard J. and Lu, Ming Y. and Zhang, Andrew and Vaidya, Anurag J. and Jaume, Guillaume and Shaban, Muhammad and Kim, Ahrong and Williamson, Drew F. K. and Robertson, Harry and Chen, Bowen and Almagro-Pérez, Cristina and Doucet, Paul and Sahai, Sharifa and Chen, Chengkuan and Chen, Christina S. and Komura, Daisuke and Kawabe, Akihiro and Ochi, Mieko and Sato, Shinya and Yokose, Tomoyuki and Miyagi, Yohei and Ishikawa, Shumpei and Gerber, Georg and Peng, Tingying and Le, Long Phi and Mahmood, Faisal},
  year    = {2025},
  pages   = {3749--3761}
}

@article{babak_diagnostic_2017,
  author  = {Ehteshami Bejnordi, Babak and Veta, Mitko and Diest, Paul and Ginneken, Bram and Karssemeijer, Nico and Litjens, Geert and van der Laak, Jeroen and Hermsen, Meyke and Manson, Quirine and Balkenhol, Maschenka and Geessink, Oscar and Stathonikos, Nikolaos and van Dijk, Marcory and Bult, Peter and Beca, Francisco and Beck, Andrew and Wang, Dayong and Khosla, Aditya and Gargeya, Rishab and Venâncio, Rui},
  year    = {2017},
  pages   = {2199-2210},
  title   = {Diagnostic Assessment of Deep Learning Algorithms for Detection of Lymph Node Metastases in Women With Breast Cancer},
  volume  = {318},
  journal = {JAMA},
}

@article{defazio_road_2024,
  title         = {The Road Less Scheduled},
  author        = {Aaron Defazio and Xingyu Yang and Harsh Mehta and Konstantin Mishchenko and Ahmed Khaled and Ashok Cutkosky},
  year          = {2024},
  journal        = {arXiv arXiv:2405.15682},
}

@inproceedings{mi_svm_andrews_2002,
  author    = {Andrews, Stuart and Tsochantaridis, Ioannis and Hofmann, Thomas},
  title     = {Support vector machines for multiple-instance learning},
  year      = {2002},
  booktitle = {Proceedings of the 16th International Conference on Neural Information Processing Systems},
  pages     = {577--584},
  series    = {NIPS'02}
}

@inproceedings{kopp2025utilizing,
  title     = {Utilizing {TabPFN} for Multi-Instance Data with Scarce Labels},
  author    = {Nikolaus Kopp and Alexander Fuchs and Markus Feuerstein and Phillip Paller and Franz Pernkopf},
  booktitle = {EurIPS 2025 Workshop: AI for Tabular Data},
  year      = {2025},
}

@article{waqas_exploring_2024,
  title    = {Exploring Multiple Instance Learning ({MIL}): A brief survey},
  author   = {Waqas, Muhammad and Ahmed, Syed Umaid and Tahir, Muhammad Atif and Wu, Jia and Qureshi, Rizwan},
  journal  = {Expert Systems with Applications},
  volume   = {250},
  pages    = {123893},
  year     = {2024},
}

@article{flanders_construction_2019,
  author  = {Flanders, Adam E. and Prevedello, Luciano M. and Shih, George and Halabi, Safwan S. and Kalpathy-Cramer, Jayashree and Ball, Robyn and Mongan, John T. and Stein, Anouk and Kitamura, Felipe C. and Lungren, Matthew P. and Choudhary, Gagandeep and Cala, Lesley and Coelho, Luiz and Mogensen, Monique and Mor\'{o}n, Fanny and Miller, Elka and Ikuta, Ichiro and Zohrabian, Vahe and McDonnell, Olivia and Lincoln, Christie and Shah, Lubdha and Joyner, David and Agarwal, Amit and Lee, Ryan K. and Nath, Jaya},
  title   = {Construction of a Machine Learning Dataset through Collaboration: The {RSNA} 2019 Brain {CT} Hemorrhage Challenge},
  journal = {Radiology: Artificial Intelligence},
  volume  = {2},
  number  = {3},
  pages   = {e190211},
  year    = {2020}
}

@article{castro_torchmil_2026,
  title   = {Torchmil: A PyTorch-based library for deep multiple instance learning},
  author  = {Castro-Mac{\'\i}as, Francisco M and S{\'a}ez-Maldonado, Francisco J and Morales-{\'A}lvarez, Pablo and Molina, Rafael},
  journal = {Neurocomputing},
  year    = {2026},
  volume  = {680},
  pages   = {133286},
}

@article{geman_neural_1992,
    author = {Geman, Stuart and Bienenstock, Elie and Doursat, Ren\'{e}},
    title = {Neural networks and the bias/variance dilemma},
    year = {1992},
    volume = {4},
    number = {1},
    journal = {Neural Computation},
    pages = {1–58},
}

@article{baldi_parameterized_2016,
  title   = {Parameterized neural networks for high-energy physics},
  volume  = {76},
  number  = {5},
  journal = {The European Physical Journal C},
  author  = {Baldi, Pierre and Cranmer, Kyle and Faucett, Taylor and Sadowski, Peter and Whiteson, Daniel},
  year    = {2016},
  pages   = {235}
}

@article{carbonneau_multiple_2018,
  title   = {Multiple instance learning: {A} survey of problem characteristics and applications},
  volume  = {77},
  journal = {Pattern Recognition},
  author  = {Carbonneau, Marc-Andr{\'e} and Cheplygina, Veronika and Granger, Eric and Gagnon, Ghyslain},
  year    = {2018},
  pages   = {329--353}
}

@article{dietterich_solving_1997,
  title   = {Solving the multiple instance problem with axis-parallel rectangles},
  volume  = {89},
  number  = {1},
  journal = {Artificial Intelligence},
  author  = {Dietterich, Thomas G. and Lathrop, Richard H. and Lozano-Pérez, Tomás},
  year    = {1997},
  pages   = {31--71}
}

@article{frey_letter_1991,
  title   = {Letter recognition using {Holland}-style adaptive classifiers},
  volume  = {6},
  number  = {2},
  journal = {Machine Learning},
  author  = {Frey, Peter W. and Slate, David J.},
  year    = {1991},
  pages   = {161--182}
}

@inproceedings{zaheer_deep_2017,
  title     = {Deep {Sets}},
  volume    = {30},
  booktitle = {Advances in {Neural} {Information} {Processing} {Systems}},
  author    = {Zaheer, Manzil and Kottur, Satwik and Ravanbakhsh, Siamak and Poczos, Barnabas and Salakhutdinov, Russ R and Smola, Alexander},
  year      = {2017},
  pages = {3394--3404}
}

@inproceedings{kolberg_tabpfn-wide_2025,
  title     = {{TabPFN}-{Wide}: Continued Pre-Training for Extreme Feature Counts},
  author    = {Kolberg, Christopher and Eggensperger, Katharina and Pfeifer, Nico},
  year      = {2025},
  booktitle = {EurIPS 2025 Workshop: AI for Tabular Data},
}

@inproceedings{garg_real-tabpfn_2025,
  title     = {Real-{TabPFN}: {Improving} {Tabular} {Foundation} {Models} via {Continued} {Pre}-training {With} {Real}-{World} {Data}},
  author    = {Garg, Anurag and Ali, Muhammad and Hollmann, Noah and Purucker, Lennart and Müller, Samuel and Hutter, Frank},
  year      = {2025},
  booktitle = {Proceedings of the 1st ICML Workshop on Foundation Models for Structured Data},
}

@article{hoo_tables_2026,
	title = {From Tables to Time: Extending {TabPFN}-v2 to Time Series Forecasting},
	author = {Hoo, Shi Bin and Müller, Samuel and Salinas, David and Hutter, Frank},
	year = {2025},
	journal = {arXiv arXiv:2501.02945},
}

@article{soklaski_tools_2022,
  title={Tools and Practices for Responsible {AI} Engineering},
  author={Soklaski, Ryan and Goodwin, Justin and Brown, Olivia and Yee, Michael and Matterer, Jason},
  journal={arXiv preprint arXiv:2201.05647},
  year={2022}
}

@article{pedregosa_scikit-learn_2011,
	title = {Scikit-learn: {Machine} {Learning} in {Python}},
	volume = {12},
	number = {85},
	journal = {Journal of Machine Learning Research},
	author = {Pedregosa, Fabian and Varoquaux, Gaël and Gramfort, Alexandre and Michel, Vincent and Thirion, Bertrand and Grisel, Olivier and Blondel, Mathieu and Prettenhofer, Peter and Weiss, Ron and Dubourg, Vincent and Vanderplas, Jake and Passos, Alexandre and Cournapeau, David and Brucher, Matthieu and Perrot, Matthieu and Duchesnay, Édouard},
	year = {2011},
	pages = {2825--2830},
}

@inproceedings{paszke_pytorch_2019,
    author = {Paszke, Adam and Gross, Sam and Massa, Francisco and Lerer, Adam and Bradbury, James and Chanan, Gregory and Killeen, Trevor and Lin, Zeming and Gimelshein, Natalia and Antiga, Luca and Desmaison, Alban and K\"{o}pf, Andreas and Yang, Edward and DeVito, Zach and Raison, Martin and Tejani, Alykhan and Chilamkurthy, Sasank and Steiner, Benoit and Fang, Lu and Bai, Junjie and Chintala, Soumith},
    title = {PyTorch: an imperative style, high-performance deep learning library},
    year = {2019},
    booktitle = {Proceedings of the 33rd International Conference on Neural Information Processing Systems},
    numpages = {12},
    pages = {8026--8037}
}

@inproceedings{pfefferle_nanotabpfn_2025,
	title = {{nanoTabPFN}: {A} {Lightweight} and {Educational} {Reimplementation} of {TabPFN}},
	author = {Pfefferle, Alexander and Hog, Johannes and Purucker, Lennart and Hutter, Frank},
    year = {2025},
    booktitle = {EurIPS 2025 Workshop: AI for Tabular Data},
}

@inproceedings{maron_mil_1997,
  title   = {A framework for multiple-instance learning},
  author  = {Maron, Oded and Lozano-P{\'e}rez, Tom{\'a}s},
  booktitle = {{Advances} in {Neural} {Information} {Processing} {Systems}},
  volume  = {10},
  year    = {1997},
  pages   = {570--576}
}

@inproceedings{ilse_attention_2018,
  title = 	 {Attention-based {Deep} {Multiple} {Instance} {Learning}},
  author =     {Ilse, Maximilian and Tomczak, Jakub and Welling, Max},
  booktitle =  {Proceedings of the 35th {International} {Conference} on {Machine} {Learning}},
  series =     {Proceedings of Machine Learning Research},
  pages = 	 {2127--2136},
  year = 	     {2018},
  volume = 	 {80},
}

@article{foulds_review_2010,
  title     = {A review of multi-instance learning assumptions},
  author    = {Foulds, James and Frank, Eibe},
  journal   = {The Knowledge Engineering Review},
  volume    = {25},
  number    = {1},
  pages     = {1--25},
  year      = {2010},
}

@inproceedings{hense_xmil_2024,
  title     = {x{MIL}: Insightful Explanations for Multiple Instance Learning in Histopathology},
  author    = {Julius Hense and Mina Jamshidi Idaji and Oliver Eberle and Thomas Schnake and Jonas Dippel and Laure Ciernik and Oliver Buchstab and Andreas Mock and Frederick Klauschen and Klaus Robert Muller},
  booktitle = {{Advances} in {Neural} {Information} {Processing} {Systems}},
  pages     = {8300--8328},
  year      = {2024}
}

@inproceedings{shao_transmil_2021,
  title   = {Trans{MIL}: Transformer based correlated multiple instance learning for whole slide image classification},
  author  = {Shao, Zhuchen and Bian, Hao and Chen, Yang and Wang, Yifeng and Zhang, Jian and Ji, Xiangyang and others},
  booktitle = {{Advances} in {Neural} {Information} {Processing} {Systems}},
  volume  = {34},
  pages   = {2136--2147},
  year    = {2021}
}

@inproceedings{zhao_predicting_2020,
  title     = {Predicting lymph node metastasis using histopathological images based on multiple instance learning with deep graph convolution},
  author    = {Zhao, Yu and Yang, Fan and Fang, Yuqi and Liu, Hailing and Zhou, Niyun and Zhang, Jun and Sun, Jiarui and Yang, Sen and Menze, Bjoern and Fan, Xinjuan and others},
  booktitle = {Proceedings of the {IEEE/CVF} conference on computer vision and pattern recognition},
  pages     = {4837--4846},
  year      = {2020}
}

@article{campanella_clinical_2019,
  title   = {Clinical-grade computational pathology using weakly supervised deep learning on whole slide images},
  author  = {Campanella, Gabriele and Hanna, Matthew G and Geneslaw, Luke and Miraflor, Allen and Werneck Krauss Silva, Vitor and Busam, Klaus J and Brogi, Edi and Reuter, Victor E and Klimstra, David S and Fuchs, Thomas J},
  journal = {Nature Medicine},
  volume  = {25},
  number  = {8},
  pages   = {1301--1309},
  year    = {2019}
}

@inproceedings{lu_visual_2023,
  author    = {Lu, Ming Y. and Chen, Bowen and Zhang, Andrew and Williamson, Drew F. K. and Chen, Richard J. and Ding, Tong and Le, Long Phi and Chuang, Yung-Sung and Mahmood, Faisal},
  title     = {Visual Language Pretrained Multiple Instance Zero-Shot Transfer for Histopathology Images},
  booktitle = {Proceedings of the {IEEE/CVF} Conference on Computer Vision and Pattern Recognition ({CVPR})},
  year      = {2023},
  pages     = {19764-19775}
}

@article{meseguer_mil-adapter_2026,
  title   = {{MIL}-{Adapter}: Coupling multiple instance learning and vision-language adapters for few-shot slide-level classification},
  journal = {Medical Image Analysis},
  volume  = {110},
  pages   = {103964},
  year    = {2026},
  author  = {Pablo Meseguer and Rocío {del Amor} and Valery Naranjo}
}

@article{dawood_confounding_2026,
	title = {Confounding factors and biases abound when predicting molecular biomarkers from histological images},
	journal = {Nature Biomedical Engineering},
	author = {Dawood, Muhammad and Branson, Kim and Tejpar, Sabine and Rajpoot, Nasir and Minhas, Fayyaz ul Amir Afsar},
	year = {2026},
	pages = {1--15},
}

@article{howard_signatures_2021,
	title = {The impact of site-specific digital histology signatures on deep learning model accuracy and bias},
	volume = {12},
	number = {1},
	journal = {Nature Communications},
	author = {Howard, Frederick M. and Dolezal, James and Kochanny, Sara and Schulte, Jefree and Chen, Heather and Heij, Lara and Huo, Dezheng and Nanda, Rita and Olopade, Olufunmilayo I. and Kather, Jakob N. and Cipriani, Nicole and Grossman, Robert L. and Pearson, Alexander T.},
	year = {2021},
	pages = {4423},
}

@article{jeong_scmild_2026,
	title = {{scMILD}: {Single}-cell multiple instance learning for sample classification and associated subpopulation discovery},
	volume = {29},
	number = {4},
	journal = {iScience},
	author = {Jeong, Kyeonghun and Choi, Jinwook and Kim, Kwangsoo},
	year = {2026},
}

@article{koemen_towards_2025,
  title   = {Towards Robust Foundation Models for Digital Pathology},
  author  = {K{\"o}men, Jonah and de Jong, Edwin D and Hense, Julius and Marienwald, Hannah and Dippel, Jonas and Naumann, Philip and Marcus, Eric and Ruff, Lukas and Alber, Maximilian and Teuwen, Jonas and others},
  journal = {arXiv preprint arXiv:2507.17845},
  year    = {2025}
}

@inproceedings{drexlin_medi_2025,
  title={Medi: Metadata-guided diffusion models for mitigating biases in tumor classification},
  author={Drexlin, David Jacob and Dippel, Jonas and Hense, Julius and Preni{\ss}l, Niklas and Montavon, Gr{\'e}goire and Klauschen, Frederick and M{\"u}ller, Klaus-Robert},
  booktitle={International Conference on Medical Image Computing and Computer-Assisted Intervention},
  pages={379--388},
  year={2025},
}

@article{idaji_beyond_2026,
    title={Beyond Attention Heatmaps: How to Get Better Explanations for Multiple Instance Learning Models in Histopathology}, 
    author={Mina Jamshidi Idaji and Julius Hense and Tom Neuhäuser and Augustin Krause and Yanqing Luo and Oliver Eberle and Thomas Schnake and Laure Ciernik and Farnoush Rezaei Jafari and Reza Vahidimajd and Jonas Dippel and Christoph Walz and Frederick Klauschen and Andreas Mock and Klaus-Robert Müller},
    journal = {Medical Image Analysis},
    year    = {2026},
    pages   = {104148},
}

@article{mollers_mind_2026,
  title={Mind the Gap: Continuous Magnification Sampling for Pathology Foundation Models},
  author={M{\"o}llers, Alexander and Hense, Julius and Schulz, Florian and Milbich, Timo and Alber, Maximilian and Ruff, Lukas},
  journal={arXiv preprint arXiv:2601.02198},
  year={2026}
}

@inproceedings{otsu_multiresolution_2023,
    author = {{Otsu}, Mitsuyoshi and {Nakamura}, Sho and {Tomita}, Shigeru and {Suhama}, Tomoyuki and {Shimazaki}, Yasunobu and {Nishimura}, Katsuya},
    title = {Multi-resolution domain adaptation via multiple instance learning for improving the recognition accuracy of Japanese oak wilt in low-resolution satellite imagery},
    booktitle = {SPIE Future Sensing Technologies 2023},
    year = {2023},
    volume = {12327},
    pages = {1232716},
}

@inproceedings{zhang_attention_2024,
  author    = {Yunlong Zhang and Honglin Li and Yuxuan Sun and Sunyi Zheng and Chenglu Zhu and Lin Yang},
  title     = {Attention-Challenging Multiple Instance Learning for Whole Slide Image Classification},
  booktitle = {Computer Vision -- {ECCV} 2024},
  series    = {Lecture Notes in Computer Science},
  volume    = {15111},
  pages     = {125--143},
  year      = {2024},
}

@inproceedings{li_dual-stream_2021,
  author    = {Bin Li and Yin Li and Kevin W. Eliceiri},
  title     = {Dual-Stream Multiple Instance Learning Network for Whole Slide Image Classification with Self-Supervised Contrastive Learning},
  booktitle = {Proceedings of the {IEEE/CVF} Conference on Computer Vision and Pattern Recognition {(CVPR)}},
  pages     = {14318--14328},
  year      = {2021},
}
\newpage

\appendix

\section*{Appendix}

\section{Learning Curves} \label{app:add_benchm}

\begin{figure}[ht]
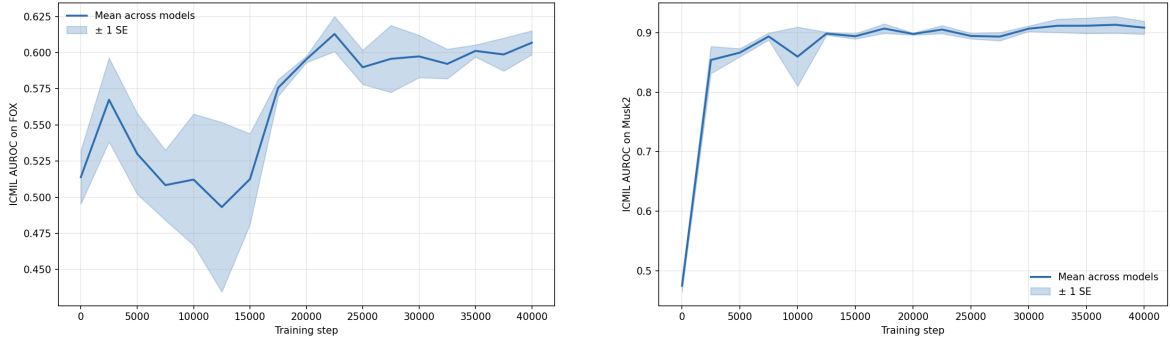

    \centering
    \begin{minipage}[t]{0.48\textwidth}
        \centering
        \includegraphics[width=\textwidth]{assets/FOX_AUROC.pdf}
    \end{minipage}
    \hfill
    \begin{minipage}[t]{0.48\textwidth}
        \centering
        \includegraphics[width=\textwidth]{assets/AUROC_Musk2.pdf}
    \end{minipage}
    \caption{Learning curves on Fox (left) and Musk2 (right), mean $\pm$ SE across three runs. Performance keeps improving with longer training, motivating the extended schedule used for ICMIL.}
    \label{fig:icmil_benchmarks}
\end{figure}

\section{Training Details}\label{app:training_details}

\paragraph{Synthetic datasets.}
Synthetic datasets are sampled from a fixed pre-generated pool. Across $20{,}000$ steps with batch size $128$, the model is exposed to approximately $2.56$M independently sampled synthetic datasets, doubling to $5.12$M for the scaled ICMIL model. No synthetic dataset is seen twice during training. Each batch is homogeneous in the sense that all $128$ datasets are drawn from the same prior configuration. Across all priors, we apply TabICL's regression-to-classification (``reg2cls'') transform~\citep{qu_tabicl_2025} to convert the SCM's continuous output into a discrete bag label.

\paragraph{Curriculum.}
To speed up training, we apply a three-stage curriculum over bag size and instance class count. In the first $500$ steps, bag sizes are restricted to $[2, 8]$ with at most $6$ instance classes. From step $500$ to $7{,}500$, bag sizes grow to $[4, 15]$ with up to $12$ instance classes. For the remainder of training (steps $7{,}500$--$20{,}000$), bag sizes reach $[6, 20]$ with up to $20$ instance classes.

\paragraph{Architecture.}
The reduced model used for the prior ablations (\S\ref{sec:exp:priors}, \S\ref{sec:exp:mixing_priors}) has $T{=}6$ iterations, $4$ attention heads, embedding dimension $E{=}128$, MLP hidden size $512$, and feature group size $s{=}1$. The scaled ICMIL model (\S\ref{sec:exp:lowdata}) keeps the same number of iterations and attention heads, increases $E$ to $256$ and the MLP hidden size to $1054$, and is trained for $40{,}000$ steps. All other architectural settings (column-row attention, chunked instance aggregation, label embedding) are identical.

\paragraph{Optimization.}
We optimize the negative log-likelihood on the query bags with the schedule-free AdamW optimizer~\citep{defazio_road_2024}. For the reduced setup used in the prior ablations (\S\ref{sec:exp:priors}, \S\ref{sec:exp:mixing_priors}) we use learning rate $1{\times}10^{-3}$ with no warmup. For the scaled ICMIL model (\S\ref{sec:exp:lowdata}) we use learning rate $5{\times}10^{-4}$ with $2{,}500$ warmup steps to stabilize training. All models are trained on a single NVIDIA A100 GPU.

\subsection{Implementation} \label{sec:implementation}

We use \texttt{PyTorch} \citep{paszke_pytorch_2019} to implement our experiments and configure them via \texttt{hydra-zen} \citep{soklaski_tools_2022}. Our synthetic priors build on the MLP-SCM and Tree-SCM generators from \texttt{TabICL} \citep{qu_tabicl_2025}, including its hyperparameter samplers and its regression-to-classification (\texttt{reg2cls}) transform. The ICMIL Perceiver-style architecture and training loop are implemented on top of the \texttt{nanoTabPFN} reference implementation \citep{pfefferle_nanotabpfn_2025}, and we optimize with schedule-free AdamW from \texttt{schedule-free} \citep{defazio_road_2024}. For baselines, we use \texttt{ABMIL} \citep{ilse_attention_2018} as re-implemented in \texttt{MIL-Lab} \citep{shao_do_2025}, and the \texttt{TabPFN-v2} model from \texttt{TabPFN} \citep{hollmann_accurate_2025}; all classical baselines (logistic regression, SVM, PCA, stratified cross-validation) use \texttt{scikit-learn} \citep{pedregosa_scikit-learn_2011}. Benchmark loaders for RSNA-ICH are taken from \texttt{torchmil} \citep{castro_torchmil_2026}.

We run our experiments on a high-performance cluster. ICMIL pretraining runs are executed on a single NVIDIA A100 with 80~GB of VRAM. Training the reduced models used in the prior ablations (\S\ref{sec:exp:priors}, \S\ref{sec:exp:mixing_priors}) takes approximately 12~h per run, while the scaled ICMIL model (\S\ref{sec:exp:lowdata}) trains for roughly 24~h. Evaluation runs for the supervised baselines (ABMIL, DSMIL, ACMIL, MeanLogReg, SVM-Summ, TabPFN-Concat, TabPFN-Subsample) are executed on less powerful GPUs of the same cluster. 

\section{Ablations} \label{app:ablations}

We isolate the contribution of the components that our architecture introduces for bag-structured inputs (\S\ref{sec:arch}), namely the iterative aggregation of instances into bag tokens and the attention between bags, and additionally probe the embedding dimension and the weighting of the prior mixture. All ablations use the reduced setup of Appendix~\ref{app:training_details} and train on the \textsc{Mixed} prior for $20{,}000$ steps with the same seed. We vary one component at a time and report mean AUROC across the twelve benchmarks. Rows marked \emph{(paper)} correspond to the configuration used in the main text.

\paragraph{Iterative aggregation and bag-token updates.}
At $T{=}1$ the model cannot refine the bag token after the labels are seen and underperforms drastically ($65.5$, Table~\ref{tab:arch_ablations}). Iterating the aggregation is therefore necessary, with performance rising steeply through $T{=}2$ ($79.0$) and stabilizing around $T{=}4$ ($83.3$).

\paragraph{Inter-bag attention.}
In-context learning is impossible without information exchange between bag representations. Disabling inter-bag attention leaves each bag token with access only to its own instances, and performance collapses to chance ($45.8$ vs.\ $83.3$, Table~\ref{tab:arch_ablations}).

\paragraph{Embedding dimension.}
Embeddings are required for the attention computation, but their size has little effect between $E{=}48$ and $E{=}128$ ($82.4$--$83.3$, Table~\ref{tab:arch_ablations}).

\begin{table}[ht]
  \centering
  \setlength{\aboverulesep}{0pt}\setlength{\belowrulesep}{0pt}
  \begin{tabular}{lc}
  \toprule
  \rowstrutc Ablation & Mean AUROC \\
  \midrule
  \multicolumn{2}{>{\columncolor{gray!20}}l}{\rowstrutc\textit{Iterations}} \\
  \rowstrutc $T{=}1$ & \sv{65.5} \\
  \rowstrutc $T{=}2$ & \sv{79.0} \\
  \rowstrutc $T{=}4$ & \sv{83.3} \\
  \rowstrutc $T{=}6$ (paper) & \sv{83.3} \\
  \midrule
  \multicolumn{2}{>{\columncolor{gray!20}}l}{\rowstrutc\textit{Inter-bag attention}} \\
  \rowstrutc Disabled & \sv{45.8} \\
  \rowstrutc Enabled (paper) & \sv{83.3} \\
  \midrule
  \multicolumn{2}{>{\columncolor{gray!20}}l}{\rowstrutc\textit{Embedding dimension}} \\
  \rowstrutc $E{=}48$ & \sv{83.1} \\
  \rowstrutc $E{=}64$ & \sv{82.4} \\
  \rowstrutc $E{=}96$ & \sv{83.3} \\
  \rowstrutc $E{=}128$ (paper) & \sv{83.3} \\
  \bottomrule
  \end{tabular}
  \caption{Ablation of the architectural components, reporting mean AUROC (\%) across the twelve benchmarks. Each block varies one component while holding the remainder at the reduced setup.}
  \label{tab:arch_ablations}
\end{table}

\paragraph{Prior mixture weights.}
The mixture weights are fixed a priori, giving most of the weight to the \textsc{Joint} prior for its strong overall performance and the remainder to the two factorized priors that cover the uncorrelated witness benchmarks. As a post-hoc check of their sensitivity, we train one model per weighting and report the results in Table~\ref{tab:mixture_ablations}. Mean AUROC varies by at most $1.5$ points across the three weightings and never collapses, so the weights are not a sensitive hyperparameter. The more uniform $0.60/0.20/0.20$ weighting is marginally ahead in this reduced setup, and adopting it at scale might improve ICMIL further.

\begin{table}[ht]
  \centering
  \setlength{\aboverulesep}{0pt}\setlength{\belowrulesep}{0pt}
  \begin{tabular}{lc}
  \toprule
  \rowstrutc Mixture weights & Mean AUROC \\
  \midrule
  \rowstrutc $0.60 / 0.20 / 0.20$ & \sv{83.5} \\
  \rowstrutc $0.70 / 0.15 / 0.15$ (paper) & \sv{83.3} \\
  \rowstrutc $0.80 / 0.10 / 0.10$ & \sv{82.0} \\
  \bottomrule
  \end{tabular}
  \caption{Sensitivity to the prior mixture weights, given as \textsc{Joint} / \textsc{Factorized}$(\text{cont},\text{MLP})$ / \textsc{Factorized}$(\text{disc},\text{lookup})$ and reporting mean AUROC (\%) across the twelve benchmarks.}
  \label{tab:mixture_ablations}
\end{table}

\section{Predefined versus In-Context Aggregation} \label{app:aggregation}

TabPFN-Cluster~\citep{kopp2025utilizing} and ICMIL both build on a pretrained in-context learner, but they differ fundamentally in how instances are aggregated into a bag representation. The aggregation strategy of TabPFN-Cluster is predefined and independent of the task labels. ICMIL instead learns to aggregate instances according to the task at hand, using the labels of the context bags to decide which instance-level information to preserve.

To illustrate the difference, we construct a simple task on which TabPFN-Cluster fails and ICMIL succeeds. We generate $150$ training and $60$ test bags of $50$ instances ($25$ features) from two Gaussians (means $\pm 2$), clip feature $24$ at $\pm 3.5$, and inject witness instances beyond $\pm 4$. A bag is positive only if it contains both witnesses. We run TabPFN-Cluster with $K{=}2$ clusters, matching the true number, and ICMIL out of the box (Table~\ref{tab:aggregation}). The predefined per-cluster mean aggregation of TabPFN-Cluster averages away the witness values that define the labels. This predefined aggregation makes it fail on this task, while ICMIL can flexibly approximate the labeling function in-context and succeed. 

\begin{table}[ht]
  \centering
  \setlength{\aboverulesep}{0pt}\setlength{\belowrulesep}{0pt}
  \begin{tabular}{lllc}
  \toprule
  \rowstrutc Method & Aggregation & Uses bag labels for Aggregation & AUROC \\
  \midrule
  \rowstrutc TabPFN-Cluster & predefined & no & \sv{54.0} \\
  \rowstrutc ICMIL & learned in-context & yes & \svb{88.0} \\
  \bottomrule
  \end{tabular}
  \caption{Predefined versus in-context aggregation on the synthetic two-witness task ($150$ training bags, $60$ test bags). AUROC (\%) on the test bags.}
  \label{tab:aggregation}
\end{table}

\section{Benchmark Details} \label{app:benchmarks}

This appendix gives the full description of the twelve MIL benchmarks summarized in Section~\ref{sec:exp:priors}. The paragraphs that follow describe each benchmark. Each benchmark comprises a fixed number of bags: Musk1 has $92$ bags and Musk2 has $102$; Elephant, Fox, and Tiger contain $200$ bags each; and the remaining benchmarks (SMIL, Pos/Neg, Adjacent Pairs, Letters, HEPMASS, RSNA-ICH, and TCGA) are subsampled to $100$ bags.

\paragraph{MNIST SMIL.} Following the xMIL benchmark \citep{hense_xmil_2024}, we draw $N_\mathrm{bags}{=}100$ bags whose size $K{\sim}\mathcal{U}\{10,20\}$ is resampled per draw. For every bag we uniformly activate a random subset of the digit classes $\{0,\dots,9\}$ and sample instances with replacement from the active set; empty activations are rejected. Each instance is represented by a $512$-dim ResNet-18 embedding reduced to $d{=}25$ dimensions with PCA fit per draw on its training bags. The witness digit is sampled uniformly per draw from $\{0,\dots,9\}$, and positive bags contain exactly $w{\sim}\mathcal{U}\{1,2\}$ copies of that digit while negative bags contain none. We generate $5$ independent draws, each with a fresh label rule and freshly sampled bags, partitioned $90{:}10$ into train and test.

\paragraph{MNIST PosNeg.} Identical sampling and features to SMIL, but labels are determined by a \emph{counting} rule: for each draw a pair of disjoint size-$3$ digit triples $(P, N)$ is sampled from a random permutation of $\{0,\dots,9\}$ (e.g.\ $P{=}\{4,6,8\}$, $N{=}\{5,7,9\}$), and a bag is positive iff $|\{i : x_i \in P\}| > |\{i : x_i \in N\}|$. Unlike SMIL, a single witness is insufficient; the label depends on the balance of positive and negative evidence across the bag.

\paragraph{MNIST AdjPairs.} Identical sampling and features to SMIL. For each dataset we sample a set of adjacent digit pairs (for example $\{(1,2),(3,4)\}$) and label a bag positive if any sampled adjacent pair is jointly present in the bag. Bag sizes are sampled in $[10,20]$ and the class balance is sampled in $[0.2,0.5]$.

\paragraph{UCI~Musk1.} The classical natural-MIL benchmark \citep{dietterich_solving_1997}: $92$ molecules (bags) whose conformations form the instances, with $166$ physicochemical features per conformation. We keep the native bag structure, zero-pad all bags to the dataset-wide maximum conformation count, and reduce features to $d{=}25$ with PCA. The task is binary musk vs.\ non-musk classification. We evaluate with $5$-fold stratified cross-validation on molecules so every bag tests exactly once.

\paragraph{UCI~Musk2.} Same sampling, feature representation, and protocol as Musk1, applied to the larger Musk2 collection of $102$ molecules ($39$ musk, $63$ non-musk) with $1$ to $1{,}044$ conformations per molecule. Bags are zero-padded to the dataset-wide maximum conformation count, features are reduced to $d{=}25$ with PCA, and we evaluate with $5$-fold stratified cross-validation on molecules.

\paragraph{UCI~Letters.} We adapt the UCI Letter Recognition dataset \citep{frey_letter_1991} ($20{,}000$ instances, $16$ features, $26$ classes) to MIL by constructing bags of size $K{=}10$ with the \emph{witness-rate} recipe of \citet{carbonneau_multiple_2018}: positive bags contain $\max(1,\lfloor 0.2 K\rfloor){=}2$ witnesses drawn from the target class (letter~\texttt{A}) and $8$ non-target background instances; negative bags contain only non-target instances. We build $N_\mathrm{bags}{=}100$ bags with balanced $50{:}50$ positive/negative fractions, no PCA (since $d{=}16{\le}25$), and $5$ stratified $90{:}10$ splits.

\paragraph{UCI~HEPMASS.} We use the HEPMASS~1000 variant \citep{baldi_parameterized_2016} ($27$ features, binary signal-vs-background). Bags are constructed with the same witness-rate recipe as Letters, with the signal class as the witness concept: $100$ bags of size $10$, balanced classes, $2$ witnesses per positive bag. Features are reduced to $d{=}25$ with PCA. Splits: $5$ stratified $90{:}10$.

\paragraph{Elephant/Fox/Tiger.} Three natural-image MIL datasets of $200$ images each, evaluated separately per animal class. Each image is a bag, the segmented regions are the instances, and each region is described by $230$ color and texture features. A bag is positive if at least one region contains the target animal. Bag size varies across images and is zero-padded to the dataset-wide maximum. We evaluate with $5$-fold stratified cross-validation on bags so every image tests exactly once.

\paragraph{RSNA-ICH.} An intracranial-hemorrhage CT benchmark from the \texttt{torchmil} \citep{castro_torchmil_2026} release of the RSNA 2019 Intracranial Hemorrhage Detection dataset \citep{flanders_construction_2019}. Each CT scan is a bag and its axial slices are instances, embedded with a pretrained ResNet-50 backbone ($2{,}048$-dim). A bag is positive if at least one slice contains any of the six hemorrhage subtypes. Rather than padding to the global maximum slice count, we follow the multi-draw protocol used for our other torchmil benchmarks: $50$ independent draws are sampled from the bundled \texttt{splits.csv} train/test partition. For each draw an instance count $n \sim \mathcal{U}\{15,30\}$ is drawn uniformly; $100$ training bags with $\geq n$ slices are sampled without replacement from the train partition and trimmed to $n$ slices each, while every qualifying test bag is kept and trimmed to $n$. PCA reduces features to $d{=}25$, refit per draw on that draw's training bags. 

\paragraph{TCGA LUAD vs.\ LUSC.} A binary lung adenocarcinoma (LUAD) vs.\ lung squamous cell carcinoma (LUSC) task built from TCGA whole-slide embeddings. Each patient contributes one bag, sampled from a single slide with one tile per patient cohort to prevent slide-level leakage; instances are $1{,}536$-dim UNI2 patch features. Class balance is intentionally imbalanced at $80$ LUAD / $20$ LUSC patients, and each bag is constructed by sampling $K{=}10$ tiles uniformly without replacement from the chosen slide (with replacement when a slide has fewer than $K$ tiles). Features are reduced to $d{=}25$ with PCA fit per split on the training bags. We use $5$ stratified $90{:}10$ train/test splits drawn over the patient pool.

\paragraph{Common interface and preprocessing.} All tasks expose the same interface, yielding per split a tuple $(\mathcal{X}_\mathrm{train}, y_\mathrm{train}, \mathcal{X}_\mathrm{test}, y_\mathrm{test})$ with bag tensors of shape $(N_\mathrm{bags}, K, d)$ and bag labels of shape $(N_\mathrm{bags},)$, where $K$ is the number of instances per bag (zero-padded when variable) and $d$ is the feature dimensionality. Musk1, Musk2, and Andrews Fox/Tiger/Elephant use $5$-fold stratified cross-validation on bags so every bag tests exactly once; RSNA-ICH uses the $50$-draw protocol described above; the remaining tasks (MNIST SMIL/PosNeg/AdjPairs, Letters, HEPMASS, TCGA) use $5$ stratified $90{:}10$ train/test splits. Variable-size bags are zero-padded to the dataset-wide maximum. Per-instance features are reduced to $d{=}25$ with PCA fit per split on the training bags, with the sole exception of Letters where the raw $d{=}16 \leq 25$ leaves PCA inactive.

\end{document}